\RequirePackage{amsmath}
\documentclass[runningheads]{llncs}
\usepackage[utf8]{inputenc}

\usepackage{amsmath}
\usepackage{amssymb}
\usepackage{amsfonts}
\usepackage{listings}
\usepackage{galois}
\usepackage{algpseudocode}
\usepackage{algorithm}
\usepackage{tikz}
\usetikzlibrary{shapes}

\newcommand{\Q}{\mathbb{Q}}

\usepackage{blindtext}
\usepackage{lipsum}
\DeclareMathOperator*{\argmax}{argmax}
\DeclareMathOperator*{\parbigwedge}{\overline{\overline{\bigwedge}}}

\newcommand{\A}{\mathcal{A}}
\newcommand{\C}{\powerset{\Q}}
\newcommand{\abs}[1]{\widehat{#1}}
\newcommand{\powerset}[1]{\raisebox{.15\baselineskip}{\Large\ensuremath{\wp}}(#1)}

\newcommand*\Let[2]{\State #1 $\gets$ #2}
\algblockdefx[ForEach]{ForEach}{EndForEach}[1]{\textbf{for each} #1 \textbf{do}}{\textbf{end for}}

\begin{document}

\title{Scaling up Memory-Efficient Formal Verification Tools for Tree Ensembles}
\author{John Törnblom\inst{1} \and Simin Nadjm-Tehrani\inst{2}}
\institute{
           {Saab AB, Bröderna Ugglas gata, SE-581 88 Linköping, Sweden\\
           \email{john.tornblom@saabgroup.com}}
           \and
           {Linköping University, SE-581 83 Linköping, Sweden\\
           \email{simin.nadjm-tehrani@liu.se}}
}

\authorrunning{J. Törnblom and S. Nadjm-Tehrani}

\maketitle

\begin{abstract}
To guarantee that machine learning models yield
outputs that are not only accurate, but also robust, recent works propose
formally verifying robustness properties of machine learning models. To be
applicable to realistic safety-critical systems, the used verification 
algorithms need to manage the combinatorial explosion resulting from vast
variations in the input domain, and be able to verify correctness properties
derived from versatile and domain-specific requirements.

In this paper, we formalise the VoTE algorithm presented earlier as a tool 
description, and extend the tool set with mechanisms for systematic scalability
studies. In particular, we show a) how the separation of property checking from
the core verification engine enables verification of versatile requirements, 
b) the scalability of the tool, both in terms of time taken for verification 
and use of memory, and c) that the algorithm has attractive properties that lend 
themselves well for massive parallelisation.

We demonstrate the application of the tool in two case studies, namely digit 
recognition and aircraft collision avoidance, where the first case study serves
to assess the resource utilisation of the tool, and the second
to assess the ability to verify versatile correctness
properties.
\end{abstract}

\section{Introduction}\label{sec:introduction}
With the exposure of a wide variety of artificial intelligence (AI) and machine
learning (ML) techniques, the volume of works that study explainability in AI
and promote methods to analyse ML-based approaches with respect to 
trustworthiness of explanations are increasing exponentially~\cite{Adadi18}. 
A growing number of surveys dealing with the notion of 
explainability~\cite{Arrieta20,Tjoa20} show that the concept has many 
dimensions, e.g., interpretability, accountability, and transparency. 
When trustworthiness is discussed in this context, one considers that 
explainability increases trust in the decisions made by intelligent systems. 
However, there are concerns about the ``risk of being forced to explain the
explanations''~\cite{Adadi18}, which calls for formalised approaches to 
quantitative assessments of explainability.

In our work, we begin with the classic premise that trustworthiness in 
safety-critical applications is of prime concern, and that methods that 
increase the confidence in the systems are worth exploring. Applying formal 
verification to show that a system exhibits its intended behaviour is promoted 
as a means of increasing its trustworthiness. When ML components are envisaged 
in a system, formal verification may additionally increase the degree of 
explainability of a system's behaviour. However, to make a meaningful 
contribution in this direction, one must first apply the essential rigour 
that is the backbone of performing comparative performance studies. The use 
of ML methods makes some of these rigorous analyses harder to perform due to 
the multiple dependencies on the data used to train models, the internal
workings of the increasingly complex and opaque ML models, the set of properties
of interest that can be defined in slightly different ways, and large system 
state spaces demanding large amounts of resources during analysis. 

This paper studies formal verification of tree ensembles, a class of ML models
considered less opaque than other ML models, thus possessing more intrinsic 
explainability. The paper is centred around the tool VoTE, that uses abstract 
interpretation to reduce the search space when verifying a given property, 
and was presented informally earlier~\cite{Tornblom19}. We aim to show that, 
in addition to building on well-known mathematical concepts (conservative 
approximations and abstraction-refinement) that make the verification algorithm
sound and complete, the tool set can be extended towards useful practical 
properties that are a result of architectural and algorithmic design decisions. 
Most importantly, that systematic performance analysis is enabled in presence 
of multiple input languages for the trained ML models. Also, verification of 
different requirements are possible due to the decoupling of the property 
checking and the core abstraction-refinement engine. These basic tool 
characteristics are exploited for systematic empirical experiments showing
its time and memory scalability, and ability to verify different types of 
requirements. The contributions of the paper are as follows\footnote{Source 
code and data is available at https://github.com/john-tornblom/VoTE-scaling-experiments}:
\begin{itemize}
    \item We describe extensions of VoTE that show how benchmarking of its 
          performance can be performed transparently by providing a 
          model-as-input translation validation mechanism, which facilitates 
          systematic comparisons with other tools in a fair setting.
    \item We use time and memory as metrics of scalability, and quantify the 
          resource efficiency of VoTE in comparison with a state of art baseline. 
          Using insights from this comparison, we then show that parallelisation
          of VoTE computations is possible with little additional effort.
    \item We illustrate that VoTE is capable of verifying properties different 
          from robustness (treated earlier~\cite{Tornblom19}) through the 
          modularity of the property checker. 
          This is done by replicating the verification of ten properties of an
          aircraft collision avoidance system previously studied in the context
          of neural networks~\cite{Katz17}.
\end{itemize}
The paper is structured as follows. Section~\ref{sec:background} introduces the 
fundamental concepts used in the paper, e.g., tree ensembles, robustness, and 
abstract transformers, and Section~\ref{sec:related-works} relates our 
contributions to earlier work. In Section~\ref{sec:extension}, we formalise 
the VoTE core algorithm, and propose a couple of extensions to VoTE and its 
analysis environment. In Section~\ref{sec:scalability}, we use two case 
studies to show the scalability and versatility of VoTE, and 
Section~\ref{sec:conclusions} concludes the paper.

\section{Background}\label{sec:background}
In this section, we introduce decision trees, and describe how they are used
as main building blocks in a more advanced class of ML models called
tree ensembles. We then define robustness against input perturbations, a
property used in Section~\ref{sec:digit-recognition} for our systematic
performance comparisons since the state of art so far is focused on this
property. Finally, the section ends with a brief description of abstraction
interpretation.
\subsection{Decision Trees}
In machine learning, decision trees are used as predictive models to capture
statistical properties of a system. Let $\{X_1, \ldots, X_k\}$ be
a partition of the $n$-dimensional input domain $\Q^n$. A decision tree is then
defined as a set of $k$ pairs $T = \{(X_1, \bar{y}_1), \ldots, (X_k, \bar{y}_k)\}$,
where $\bar{y}_i$ is an $m$-dimensional value from the output domain $\Q^m$,
and a prediction function $t: \Q^n \rightarrow \Q^m$ that maps values from the
input domain to values from the output domain, i.e., 
\begin{equation}
\label{eq:decision-tree}
  t(\bar{x}) = 
  \begin{cases} 
    \bar{y}_1 & \bar{x} \in X_{1}, \\
    \hfill  & \vdots\\
    \bar{y}_k & \bar{x} \in X_{k}. \\
  \end{cases}
\end{equation}

\subsection{Tree Ensembles}
Decision trees are known to suffer from overfitting, i.e.,
the model becomes too specialized towards training data, and the prediction
function generalizes poorly when confronted by previously unseen inputs.
To counteract this phenomenon, several types of tree ensembles have been
proposed, e.g., random forests~\cite{Breiman01} and gradient boosting
machines~\cite{Friedman01}. The different types of tree ensembles are normally
distinguished by the used learning algorithms, while their prediction functions
have similar structure. Let $F = \{T_1, \ldots, T_B\}$ be a set of $B$ decision
trees. The prediction function of a tree ensemble is then defined as
\begin{equation}
  \label{eq:tree-ensemble}
  f(\bar{x}) = p\Big(\sum_{i=1}^{B} t_i(\bar{x})\Big),
\end{equation}
where $p$ is a post-processing function. In random forests, the post-processing
function divides the sum of trees with the number of trees, while gradient
boosting machines trained on classification problems use the softmax function
to post-process the sum of trees.

\subsection{Classifiers}
When tree ensembles are trained to predict probabilities, they can be combined 
with the argmax function to form a classifier. 
Let $f(\bar{x}) = (y_1, \ldots, y_m)$ be a tree ensemble trained to predict the
probability $y_i$ that a given input maps to a class $i$, where $m$ 
is the number of classes. A classifier $f_c$ is then defined as
\begin{equation}
  \label{eq:classifier}
  f_c(\bar{x}) = \argmax_{i \in  \{1, \ldots, m\}} y_i.
\end{equation}

\subsection{Robustness against Input Perturbations}
\label{sec:robustness}
In the context of machine learning, robustness describes the ability of a system
to maintain the correct prediction despite noisy or adversarial input.
Let $f_c$ be a classifier subject to verification, $X \subset \Q^n$ a set of 
input samples with label $l \in \{1, \ldots, m\}$,
$\epsilon \in \Q_{\ge 0}$ a robustness margin, and
$\Delta = \{ \delta \in \Q: -\epsilon < \delta < \epsilon \}$ a set of
possible input perturbations. We denote by $\bar{\delta}$ an $n$-tuple of
elements drawn from $\Delta$. The classifier $f_c$ is robust with respect to
$X$ and $\epsilon$ iff
\begin{equation}
  \label{eq:robustness}
  \begin{aligned}
    \forall \bar{x} \in X,
    \forall \bar{\delta} \in \Delta^n, \\
    f_c(\bar{x}) = f_c(\bar{x} + \bar{\delta}) = l.
  \end{aligned}
\end{equation}

\subsection{Abstract Interpretation}
Abstract interpretation is a framework to facilitate sound and efficient
reasoning about programs being analysed by a compiler~\cite{Cousot77}. The
idea is to transform the source code of a program that computes values in a
concrete domain into functions that operate in one or more abstract domains 
where some analyses of interest are faster than in the concrete domain, but
potentially less precise. 

For example, an abstraction function $\alpha: \C \rightarrow \A$ is used to map
a set of rational numbers $X \subseteq \Q$ to an abstract element
$\abs{x} \in \A$, where $\C$ denotes the power set over $\Q$. Analogously,
a concretisation function $\gamma: \A \rightarrow \C$ is used to map an
abstract value $\abs{y} \in \A$ to a set of rational numbers $Y \subseteq \Q$. 
Abstraction and concretisation mappings that have a certain property called
\textit{Galois connection} lead to sound reasoning with abstract 
interpretation~\cite{Cousot77}, i.e.,
\begin{equation}
\label{eq:galois-connection}
  \begin{aligned}
    \forall X \in \C, X \subseteq \gamma(\alpha(X)), \\
    \forall \abs{x} \in \A, \abs{x} = \alpha(\gamma(\abs{x})).
  \end{aligned}
\end{equation}
To perform the analysis, the program is interpreted in the abstract domain by
evaluating sequences of abstract values, operators, and \textit{transformers}.
An abstract transformer $\abs{f}: \A \rightarrow \A$ is conservative with
respect to a concrete function $f: \Q \rightarrow \Q$ iff
\begin{equation}
  \label{eq:conservative-transformation}
  \begin{aligned}
    \forall X \in \powerset{\Q}, 
    \forall x \in X, \\
    f(x) \in \gamma(\abs{f}(\alpha(X))).
  \end{aligned}
\end{equation}
An example operator on the abstract domain we use in Section~\ref{sec:extension}
is the join operator $\sqcap$, which is conservative with respect to the
set intersection operator $\cap$, i.e., 
$\gamma(\abs{x}_1) \cap \gamma(\abs{x}_2) \subseteq \gamma(\abs{x}_1 \sqcap \abs{x}_2)$.

\section{Related Works}
\label{sec:related-works}
The formal methods community has a long tradition of rigorous and systematic 
tool evaluations, often arranged in the form of competitions~\cite{Bartocci19}.
These arrangements help to ensure that tools are evaluated on a fair and 
transparent basis, and that experiments are reproducible. Unfortunately, there
is no well-established competition for formal verification of machine learning
models. However, Nicolae et al.~\cite{Nicolae18} present the Adversarial 
Robustness Toolbox, a library devoted to the evaluation of different learning
and verification algorithms in the context of robustness. In this paper, we 
broaden their scope beyond the robustness property.

Liu et al.~\cite{Liu21} present a comparison between many different
verification algorithms designed specifically for neural networks, as well as
evaluating their run-time performance when verifying robustness in a digit
recognition case study~\cite{MNIST}, and domain-specific requirements of an
aircraft collision avoidance system~\cite{Katz17}. In this paper, we provide
the means to perform transparent and systematic evaluations for tools designed
specifically to verify tree ensembles. We reuse the above two case studies, 
but with some additional work. In particular, we verify robustness of the 
entire test set in the digit recognition case study, as opposed to a single
image as was done in the named study. Since the data used to train the 
aircraft collision avoidance system is not publicly available (but only the
resulting neural network), we use the published models as an oracle to sample 
data that we use to train tree ensembles. We then verify the properties of the
created tree ensembles using the same domain-specific properties originally
suggested by Katz et al.~\cite{Katz17} to verify neural networks.

For tree ensembles, only a few formal verification tools have been published.
Our own earlier work presents the tool VoTE with some heuristic studies, but no 
comparative scalability analyses~\cite{Tornblom20}. Later work~\cite{Tornblom19}
presents the adaptation of the tool with abstraction-refinement, with an 
emphasis on robustness verification. In this paper, we formalise the algorithm
used in the tool engine, and extend the tool suite with means for systematic 
studies and rigorous comparisons with related state-of-art. 
Analysis of memory complexity is exploited for parallel executions that show
memory efficiency results and timing improvements. Moreover, we extend the set
of properties verified to show the versatility of the tool's property checking 
component.

Einziger et al.~\cite{Einziger19} present the tool VeriGB for verifying the
robustness of gradient boosting machines. They encode the verification problem
as an SMT formula, and use an SMT solver for verification. They design the SMT
formula to facilitate parallel analyses for different counter-examples. 
Similarly, Davos et~al.~\cite{Devos21} also use an SMT solver to verify tree
ensembles, but partition the input space into regions which are analysed in
parallel. In this paper, we take a similar approach to parallelism as 
Davos~et~al.

Ranzato and Zanella~\cite{Ranzato20} present a tool called Silva, which is based
on a similar abstraction-refinement approach as proposed in~\cite{Tornblom19}. 
Their paper is, to the best of our knowledge, the only work where run-time 
performance of several verifiers for tree ensembles is assessed. However, their
work did not include memory efficiency as a metric. Also, their analysis of 
timing did not compare verification on same artefacts (there were differences 
in models and timeouts), and used different metrics for comparative tools. This
is rectified in our work with the extension to the VoTE tool suite so that 
systematic comparisons can be performed.

\section{Formalising and Extending VoTE}
\label{sec:extension}
In this section, we begin by providing an algorithmic overview of VoTE that
formalises the earlier tool description~\cite{Tornblom19} in the abstract
interpretation framework. This helps us to understand the empirical results 
from scalability assessments conducted later in Section~\ref{sec:digit-recognition}. 
We then go on to propose a couple of extensions to VoTE and its analysis 
environment, which facilitate systematic scalability analyses (with alternative 
tools plugged in), and the ability to parallelise the analyses across multiple
CPU cores.

\subsection{The VoTE Algorithm}
Let $\abs{f}: \A^n \rightarrow \A^m$ be a
transformer that is conservative with respect to the tree ensemble prediction
function $f$. The VoTE algorithm can then be characterised as a Boolean 
function as defined by Algorithm~\ref{algo:VoTE}, 
where $F = \{T_1, \ldots, T_B\}$ is the set of trees in the ensemble,
$T_i = \{(X_1, \bar{y}_1), \ldots, (X_k, \bar{y}_k)\}$ the $i$-th tree in the
ensemble, $\abs{x}$ an abstracted input region, 
$\powerset{\Q^n} \galois{\alpha}{\gamma} \A^n$ a Galois connection, 
and $c$ a \textit{property checker}.
The property checker ensures that concrete input values captured by the abstract
input region $\abs{x}$ are in the relevant relation to concrete output values 
captured by an abstract output region $\abs{y}$. It either returns $Pass$ when 
the inputs and outputs are in the relevant relationship, $Fail$ when the absence
of the desired relationship can be proved, and $Unsure$ when this cannot be
proved~\cite{Tornblom19}.

\begin{algorithm}
  \caption{The Boolean verification algorithm implemented in VoTE, which takes
    as input a set of trees ($F$), an abstract input region ($\abs{x}$), and a
    property checker ($c$).}
  \label{algo:VoTE}
  \begin{algorithmic}[1]
    \Function{VoTE}{$F, \abs{x}, c$}
      \Let{$\abs{y}$}{$\abs{f}(\abs{x})$}
      \Comment{Interpret tree ensemble}
      \Let{$o$}{$c(\abs{x}, \abs{y})$}
      \Comment{Check the property}
      \If{$o = Pass$}
        \State \Return $True$
      \ElsIf{$o = Fail$}
        \State \Return {$False$}
      \EndIf 
      \Comment{Property checker is unsure}
      \State {$T \in F$}
      \Comment{Select a tree}
      \State \Return {$\bigwedge\limits_{(X_i, \bar{y}_i) \in T}\text{VoTE}(F \backslash \{T\}, \abs{x} \sqcap \alpha(X_i), c)$}
      \Comment{Refine the abstraction}
    \EndFunction
  \end{algorithmic}
\end{algorithm}

Given the input region $\abs{x}$ defined by the abstraction function
$\alpha$ applied to a set of values from the input domain of a model, the VoTE 
algorithm uses the transformer $\abs{f}$ to compute the abstract output
region $\abs{y}$ that approximates the image of the model for the given
input region (line 2). Next, the algorithm invokes the property checker, which 
checks if the property is satisfied for the given inputs and outputs. 
If the property checking is inconclusive, the algorithm selects an arbitrary 
tree $T$ from the set $F$, and uses the removed tree's input partition 
$\{X_1, \ldots, X_k\}$ (see Equation~\ref{eq:decision-tree}) to refine $\abs{x}$
into smaller input regions (line 10). 
This procedure is then recursively applied to smaller and smaller input regions, 
ending with a precise abstraction of the output when the set $F$ is empty. In
other words, the right hand side of Equation~\ref{eq:conservative-transformation}
returns a singleton, being the lowest level of abstraction.

The algorithm does not require any particular order in the selection of trees 
(line 9). Indeed, different selection orders can influence the time  taken for
verifying the model. In the implementation of VoTE, trees are selected in the 
order that they appear in the serialised model given as input to the tool. 
This is a recurring question for many tools with heuristics governed by
internal choices. In this case, the order can be changed by simply rearranging
the trees in the serialised model given as input to the tool.
Similarly, the order of selection of regions to refine ($X_i$ in line 10) is 
arbitrary in the algorithm. In the actual implementation of VoTE, a heuristic
is applied (and has been described earlier~\cite{Tornblom20}).

Note that the parameterisation of the property checker makes it easy to
plug-and-play properties without modifying the core engine, an 
important capability when applying formal verification to realistic case studies
with diverse requirements. We also note that the algorithm is recursive with a
maximum recursion depth equal to the number of trees in the ensemble, meaning 
that the space complexity grows linearly with respect to the number of trees.

\subsection{Extensions to VoTE}
\label{sec:extension-sub}
For a rigorous performance evaluation of a verification tool, we need to 
benchmark its scalability and verification outcomes in comparison with relevant
baselines using relevant metrics. This can be challenging when tools use
different formats for their input. To address this, we develop a 
translation validation scheme that validates the correctness of format 
conversions. Let $F$ be a set of trees serialised in an unsupported format, 
$conv$ a function that translates $F$ into a supported format, and its 
inverse being $conv^{-1}$. We can then validate that $conv$ translates
$F$ correctly by checking that $F = conv^{-1}(conv(F))$. 
Hence to enable systematic evaluations of scalability, a new module has been 
added to the tool set for such input validations in a new benchmarking context. 
In Section~\ref{sec:digit-recognition}, we illustrate this with a state-of-art
tool (Silva) that uses a different input format compared to VoTE. 

Given the attractive space complexity, we note that the VoTE algorithm can 
benefit from massive parallelisation. In particular, we are able to partition 
the set of inputs relevant to a property as a set $P$ with disjoint abstractions
$\abs{x}_i$. We can then invoke VoTE in parallel for each element in $P$, 
leading to the same outcome as a sequence of invocations to VoTE, but 
potentially faster. 

For example, when verifying robustness as defined by Equation~\ref{eq:robustness}, 
we create a set $P$ with abstractions of hyperrectangles centred around points 
$\bar{x}_i \in X$, and then invoke the VoTE algorithm in parallel for each 
abstracted hyperrectangle and the robustness property checker $c_{robust}$ as 
defined by Algorithm~\ref{algo:VoTE-robustness}, where $\parbigwedge$ denotes 
parallel invocations.

\begin{algorithm}
  \caption{The parallel robustness verification algorithm, leveraging VoTE 
           as its core verification engine.}
  \label{algo:VoTE-robustness}
  \begin{algorithmic}[1]
    \Function{$||\text{IsRobust}$}{$F, X, \epsilon$}
      \Let{$\Delta$}{$\{ \delta \in \Q: -\epsilon < \delta < \epsilon \}$}
      \Comment{Set of possible input perturbations}
      \Let{$P$}{$\Big\{\alpha(\{{\bar{x}_i + \bar{\delta}} \}): \bar{\delta} \in \Delta^n, \bar{x}_i \in X \Big\}$}
      \Comment{Set of disjoint abstractions}
      \State \Return {$\parbigwedge\limits_{\abs{x}_i \in P}\text{VoTE}(F, \abs{x}_i, c_{robust})$}
      \Comment{Parallel invocations to VoTE}
    \EndFunction
  \end{algorithmic}
\end{algorithm}

\section{Scalability studies}
\label{sec:scalability}
In this section, we use two case studies to show the scalability and versatility
of VoTE. Scalability is shown both with respect to time and memory used for 
verification. For systematic benchmarking, we exploit the extensions in 
Section~\ref{sec:extension} to measure verification time in presence of the 
presented input validation scheme. This serves to bring comparable algorithms 
and relevant case studies to run on an equal footing. 
The case studies are from earlier works, a digit recognition system 
(Section~\ref{sec:digit-recognition} and \ref{sec:parallel}), and an aircraft 
collision avoidance system (Section~\ref{sec:acasxu}). The baseline used to 
illustrate our input validation scheme is the recently published verifier 
Silva~\cite{Ranzato20}. We use a compute cluster running CentOS 7.8.2003, where 
each node is equipped with an Intel Xeon Gold 6130 CPU with 32 cores, and up to 
384\,GiB RAM.

\subsection{Digit Recognition Case Study}
\label{sec:digit-recognition}
\begin{figure}[ht]
  \centering
  \includegraphics[width=7cm]{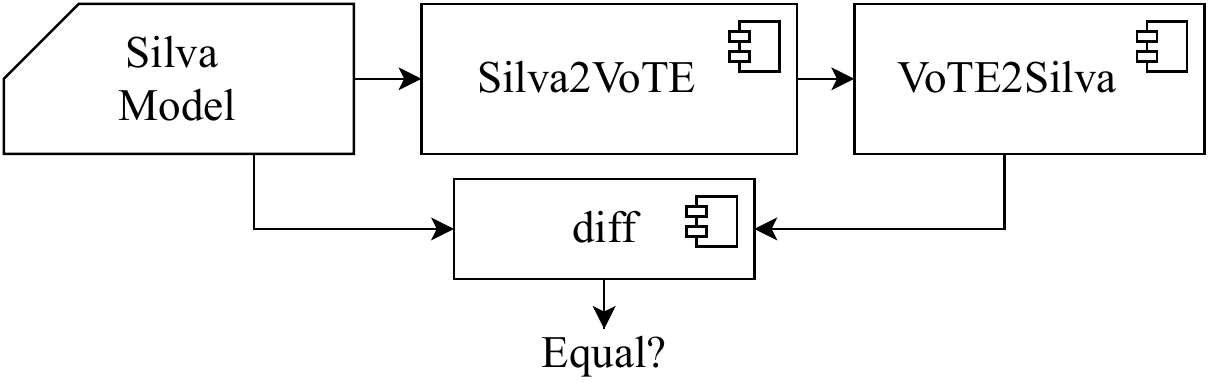}
  \caption{Our translation validation scheme applied to check equivalence of 
           VoTE and Silva input models.}
  \label{fig:translation-validation}
\end{figure}

This case study serves to assess the scalability of VoTE in terms of computational
resources when verifying the robustness of models trained on the MNIST 
dataset~\cite{MNIST}. The dataset contains 70,000 grey-scale images of 
hand-written digits with a resolution of $28 \times 28$ pixels at 8\,bpp, split
into a 85\,\% training set and a 15\,\% test set.
Normally, the intention is to use the training set together with different
learning algorithms and parameters to synthesise models. To be
comparable with related work, however, we reuse models trained by
Ranzato and Zanella~\cite{Ranzato20} in their evaluation of Silva. In particular,
we reuse gradient boosting machines trained using CatBoost~\cite{Prokhorenkova18}
with the MultiClass loss function, and random forests trained using
scikit-learn~\cite{Pedregosa11} with the Gini impurity splitting criterion. 
These models are fed to our input format conversion tool 
Silva2VoTE, and validated using the translation validation scheme proposed in 
Section~\ref{sec:extension-sub} to ensure correctness of the translations, as
illustrated by Figure~\ref{fig:translation-validation}.

While applying the scheme, we notice that leaf values in VoTE-models are 
associated with 64\,bit floating-point tuples, while leaves in Silva-models are
associated with 32\,bit integer tuples, which are normalised into 64\,bit 
floating-point tuples during analysis. Consequently, we add leaf normalisation
to the VoTE input chain so both tools can interpret the exact same input in the
same manner. With these careful configurations in the experimental setup, we 
execute the translation validation system for each model, and observe no 
discrepancies.

\subsubsection{Scalability metrics.}
Silva and VoTE both report the total elapsed time taken during verification, but
use different measurement techniques. To bring them on the same footing, we use
the GNU \verb|time| command to launch experiments and measure the elapsed wall
time. Furthermore, we use GNU \verb|time| to measure the maximum resident memory
during each experiment, i.e., the maximum amount of memory allocated on RAM 
(which excludes swap memory).

\subsubsection{Tool parameters.}
When evaluating robustness, both Silva and VoTE are parameterised with a 
robustness margin (denoted $\epsilon$ in Equation~\ref{eq:robustness}). 
In addition, Silva is parameterised with a timeout which limits the amount of 
CPU time the tool spends on verifying the robustness on a particular image. 
To evaluate both tools on a fair basis, we ensure that VoTE analyses are stopped
after the same timeout (as used in the earlier evaluation of Silva).
To sum up, we use the same parametric values as previous
works~\cite{Tornblom19,Ranzato20}, i.e., $\epsilon = 1$, and a timeout of
60 seconds per image. All experiments are executed using the version of Silva
with the git hash 9db65e58, and VoTE tagged as version 0.2.1.

\subsubsection{Scalability outcomes on a single core.}
Table~\ref{tbl:mnist} lists the elapsed time, peak memory consumption, and
number of unsolved images (due to timeouts), when Silva and VoTE are used to 
verify the robustness of random forests (RF) and gradient boosting machines 
(GBM) with different number of trees (B) and tree depths (d) on a single CPU core. 
\begin{table}
  \centering
  \caption{Verification of tree ensembles using Silva and VoTE on a single core.}
  \label{tbl:mnist}
  \small
  \begin{tabular}{|c|r|r|r|r|r|r|r|r|r|r|r|r|r|r|r|r|r}
    \hline
    \multicolumn{1}{|c|}{Model}            &
    \multicolumn{2}{c|}{Model size}        &
    \multicolumn{2}{c|}{Elapsed time\,(s)} &
    \multicolumn{2}{c|}{Memory\,(MB)}      &
    \multicolumn{2}{c|}{Unsolved}          \\ 
    \cline{2-9}
    type & B & d & Silva & VoTE & Silva & VoTE & Silva & VoTE \\
    \hline
    RF & 25 & 5 & 7 & \textbf{3} & 254 & \textbf{65} & 0 & 0 \\
    RF & 25 & 10 & 51 & \textbf{21} & 4,479 & \textbf{96} & 0 & 0 \\
    RF & 50 & 5 & 1,680 & \textbf{1,234} & 52,745 & \textbf{67} & 9 & \textbf{6} \\
    RF & 50 & 10 & 6,664 & \textbf{4,101} & 51,344 & \textbf{127} & 33 & \textbf{20} \\
    RF & 75 & 5 & 9,710 & \textbf{7,860} & 51,328 & \textbf{68} & 91 & \textbf{84} \\
    GBM & 50 & 10 & 2 & 2 & \textbf{81} & 141 & 0 & 0 \\
    GBM & 75 & 5 & 2 & 2 & 69 & \textbf{68}  & 0 & 0 \\
    GBM & 75 & 10 & 10 & \textbf{8} & 823 & \textbf{117} & 0 & 0 \\
    GBM & 100 & 10 & 136 & \textbf{83} & 9,591 & \textbf{155} & 0 & 0 \\
    GBM & 150 & 10 & 7,745 & \textbf{6,795} & 34,589 & \textbf{231} & 73 & \textbf{68} \\
    \hline
  \end{tabular}
\end{table}

Contrary to measures from earlier work~\cite{Ranzato20} (which were not carried
out in the systematic manner as we describe above), we observe that VoTE is 
typically faster in this case study, and with one exception, consumes less 
memory than Silva, especially when verifying the larger models. In the most 
time consuming experiment, VoTE is 1.2 times faster, and consumes 755 times 
less memory than Silva. Furthermore, Silva times out on a few more images than 
VoTE (as shown in the Unsolved column). In those cases where Silva and VoTE 
solve the same number of images, both tools report the same number of robust 
samples. 

\subsection{Parallelised Robustness Verification}
\label{sec:parallel}
We next run the same experiments with the parallel algorithm proposed in 
Section~\ref{sec:extension-sub}, now leveraging 32 cores. Due to the large 
memory consumption observed when verifying robustness with Silva in the 
previous section, in this section we restrict the experiments to VoTE only. 
Table~\ref{tbl:mnist-mc} lists the elapsed time and peak memory consumption of 
VoTE when utilising a single CPU core vs. using 32 CPU cores, presented in the 
same format as before. 

\begin{table}
  \centering
  \small
  \caption{Verification of tree ensembles using VoTE, executed on one and 32 cores.}
  \label{tbl:mnist-mc}
  \begin{tabular}{|c|r|r|r|r|r|r|r|r|}
    \hline
    \multicolumn{1}{|c|}{Model}            &
    \multicolumn{2}{c|}{Model size}        &
    \multicolumn{2}{c|}{Elapsed time\,(s)} &
    \multicolumn{2}{c|}{Memory\,(MB)}      &
    \multicolumn{2}{c|}{Unsolved}   \\
    \cline{2-9}
    type & B   & d  & 1\,core & 32\,cores    & 1\,core      & 32\,cores & 1\,core     & 32\,cores \\
    \hline
    RF   & 25  & 5  & 3       & 1            & \textbf{65}  & 82        & 0           & 0         \\
    RF   & 25  & 10 & 21      & \textbf{4}   & \textbf{96}  & 117       & 0           & 0         \\
    RF   & 50  & 5  & 1,234   & \textbf{84}  & \textbf{67}  & 95        & \textbf{6}  & 13        \\
    RF   & 50  & 10 & 4,101   & \textbf{249} & \textbf{127} & 166       & \textbf{20} & 36        \\
    RF   & 75  & 5  & 7,860   & \textbf{296} & \textbf{68}  & 104       & \textbf{84} & 95        \\      
    GBM  & 50  & 10 & 2       & 1            & \textbf{141} & 153       & 0           & 0         \\
    GBM  & 75  & 5  & 2       & 1            & \textbf{68}  & 76        & 0           & 0         \\
    GBM  & 75  & 10 & 8       & \textbf{2}   & \textbf{117} & 130       & 0           & 0         \\
    GBM  & 100 & 10 & 83      & \textbf{13}  & \textbf{155} & 178       & 0           & 0         \\
    GBM  & 150 & 10 & 6,795   & \textbf{282} & \textbf{231} & 267       & \textbf{68} & 79        \\
    \hline
  \end{tabular}
\end{table}

When VoTE runs analyses in parallel, we observe near-linear speedups with 
respect to CPU core count. However, we also observe time-outs on a few more
images compared to the single-core experiments, which we believe are caused by
inter-core interference. In the most time consuming experiment, VoTE with
multi-core capabilities is 26.5 times faster, consumes 1.5 times more memory,
and times out on 11 more images compared to the same single core experiment.

\subsection{Aircraft Collision Avoidance Case Study}
\label{sec:acasxu}
We now turn our attention to an early prototype of an aircraft collision
avoidance system called ACAS Xu from earlier works~\cite{Katz17}. This case
study has more diverse and application-specific properties than the robustness
property studied in the digit recognition case study. It thus highlights
the flexibility of VoTE  when confronted with more diverse requirements.

\subsubsection{System description.}
The ACAS Xu system is in part developed using a dynamic programming process that
yields a large lookup table. Each entry in the table maps state variables such
as speed and distance between vehicles to costs associated with different
actions. The action with the lowest cost is then given to the vehicle operator
as an advisory, and can be one of the following: clear of conflict, turn weak left,
turn weak right, turn strong left, or turn strong right. Since the lookup table
is very large, the mappings are compressed using a ML model. In 
previous works~\cite{Julian19a}, Julian et al. investigate the use of neural
networks for the purpose of table compression. In what follows, we use gradient
boosting machines in the same context.

Katz et al.~~\cite{Katz17} define ten properties ($\phi_{1-10}$) that the ACAS
Xu system shall satisfy. For example, property $\phi_8$ states that, for
large vertical separation and a previous weak left advisory, the system shall
either output clear of conflict, or continue advising weak left. These ten
properties alone are not sufficient for a complete safety argument, but are mere
sanity checks that ought to hold. In this paper, we use these properties solely
to illustrate VoTE versatility.

\subsubsection{Dataset.}
Unfortunately, the data used to train the ACAS Xu system 
by Julian et al.~\cite{Julian19a} and later used by Katz et al.~\cite{Katz17}
is not publicly available. Only the implementation in the form of 45 neural
networks are shared, each responsible for providing action scores in disjoint
input regions. Subsequent work by Julian and Kochenderfer~\cite{Julian19b}
include published data for a simplified variant of the ACAS Xu system, but 
without a formal specification of the requirements. Rather than adapting 
$\phi_{1-10}$ to the simplified system, we use the published neural networks as
an oracle to sample data. In particular, we sample $2 \cdot 10^6$ input
tuples uniformly across each disjoint input region, and execute the 
corresponding neural network for each sample to obtain an advisory. Since the
neural networks are known to violate some of the intended requirements, we test
each input/output pair against the available specification ($\phi_{1-10}$), and
resample any input-output pairs that violate some requirement. Hence, we obtain 
training data that we know will satisfy the requirement. Our goal is to show 
that a formal verifier will discover violations of requirements on any data on 
which the model is not trained.

\subsubsection{Systematic model synthesis.}
We train gradient boosting classifiers with 200 trees of depth 10 on 50\% of the 
sampled data using CatBoost. We set the learning rate to $0.5$, and leave the 
remaining training parameters at their default value. We then validate the model 
to observe the accuracy of the classifiers on the remaining 50\% of the data. 
This shows accuracies above 0.95 on all trained models. Since the CatBoost 
classification algorithm uses the argmax function to select a label, and the 
specification associates the minimal score with the best action, we negate the 
leaf values associated with each tree after training. We then validate that the 
trained models behave similarly to the oracle by plotting advisories for 
different downranges (distance in the direction of the vehicle) and crossranges 
(distance perpendicular to the direction of the vehicle).

\begin{figure}[ht]
  \centering
  \includegraphics[width=7cm]{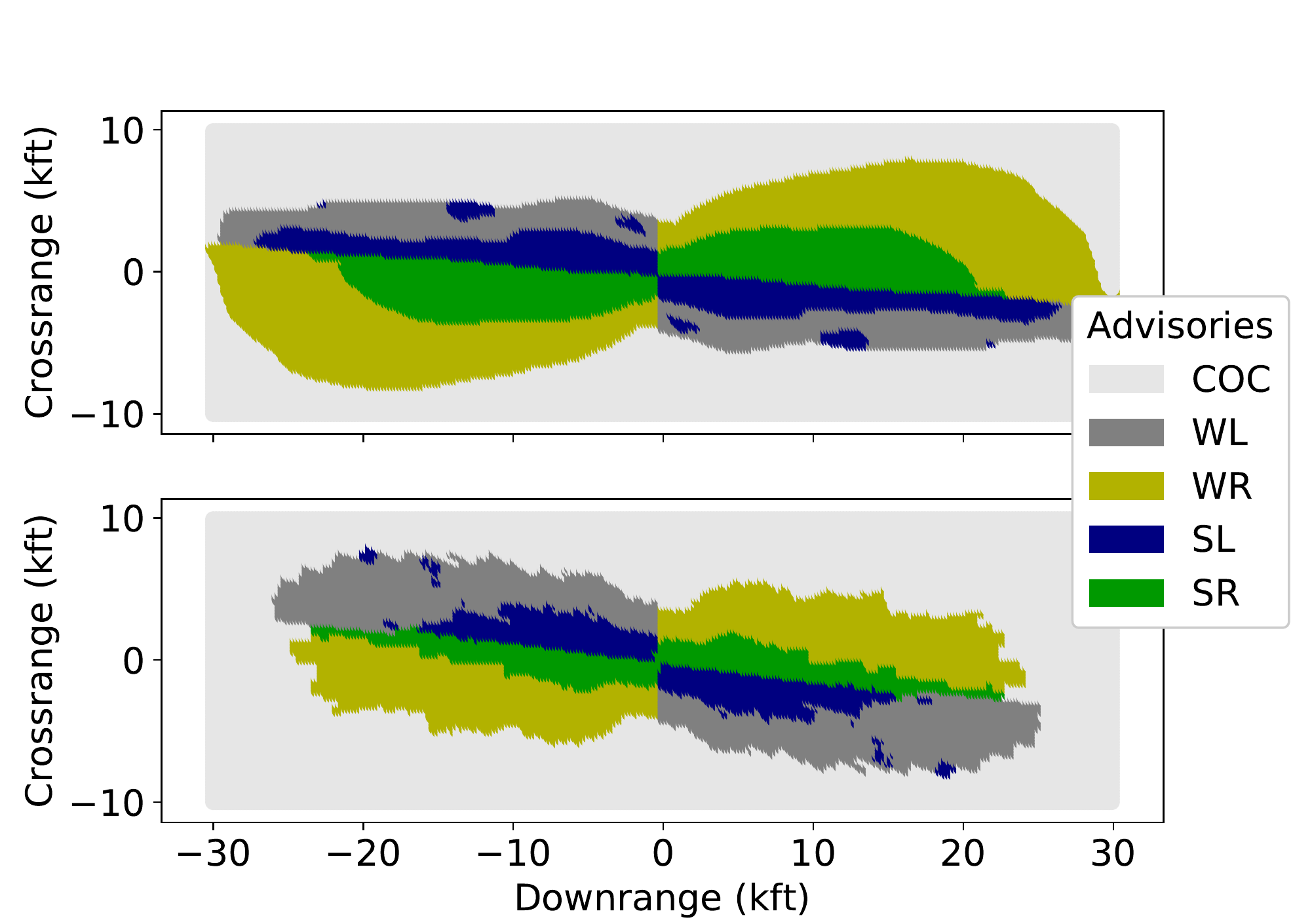}
  \caption{Advisories provided by the original neural network (top) and the
    gradient boosting machine (bottom).}
  \label{fig:acasxu}
\end{figure}

Figure~\ref{fig:acasxu} depicts the different advisories given by the oracle
and one of the trained gradient boosting machines, i.e., clear of conflict (COC),
turn weak left (WL), turn weak right (WR), turn strong left (SL), or turn strong
right (SR).

We observe that the space captured by our sample-based model captures more or 
less the same behaviour as the original neural network. Thus, we can expect
the same requirements to be valid for this system.

\subsubsection{Requirement verification.}
We execute the verification of all trained models using the same equipment as
in Section~\ref{sec:digit-recognition} with single core. Table~\ref{tbl:acasxu} 
lists the number of models that passed/failed each property, and the elapsed 
time and peak memory consumption measured using the GNU \verb|time| command.
\begin{table}
  \centering
  \caption{Verification outcome for up to 45 gradient boosting machines of ten
           properties in the ACAS Xu case study, executed on a single CPU core.}
  \label{tbl:acasxu}
  \small
  \begin{tabular}{|r|c|c|c|c|c|c|c|c|c|c|}
  \hline
  \textbf{Property}          & $\phi_1$ & $\phi_2$ & $\phi_3$ & $\phi_4$ & $\phi_5$ & $\phi_6$ & $\phi_7$ & $\phi_8$ & $\phi_9$ & $\phi_{10}$ \\ \hline
  \textbf{Pass}              & 45 & 36 & 42 & 42 & 1 & 0 & 0 & 1 & 0 & 1 \\ \hline
  \textbf{Fail}              & 0 & 4 & 0 & 0 & 0 & 1 & 1 & 0 & 1 & 0 \\ \hline
  \textbf{Elapsed time\,(s)} & 35 & 24 & 26 & 26 & 1 & 1 & 26 & 634 & 1 & 1 \\ \hline
  \textbf{Memory\,(MB)}      & 208 & 208 & 208 & 208 & 209 & 209 & 187 & 208 & 208 & 208 \\ \hline
\end{tabular}
\end{table}

Most properties take on average about a second per applicable model to verify,
with the exception of property $\phi_7$ and $\phi_8$, which takes about half
a minute and 10 minutes to verify, respectively.

\section{Conclusions}
\label{sec:conclusions}
This paper has begun the trajectory towards systematic and rigorous performance
evaluation of formal verifiers for tree ensembles, bringing both time and memory
into focus. Our work indicates that a formal underpinning for verifiers adds to
trustworthiness of the verification outcomes and ultimately the transparency of 
the deployed models. 

Our study of two of the latest tools for tree ensembles on equivalent models and 
similar metrics and running conditions helps to shed light on the scalability of
the tools, both in terms of time and memory. Our extension of VoTE with the 
translation validation scheme for an input model is a generic mechanism that can
be applied to all models used across various tools when launching rigorous 
performance studies. This work also shows that the memory footprint is a highly
relevant scalability metric for model verifiers, and a key to parallelisation
as demonstrated with the near-linear speedup in VoTE's context.

While the ACAS Xu case study demonstrates that architectural design decisions 
in VoTE affect its ability to verify diverse requirements, clearly more work 
is needed to see whether this holds in new domains with new requirements. On 
the algorithmic front, more extensive studies with variations of the heuristics 
used for selection of refinement orders would be useful.

\subsection*{Acknowledgements}
This work was partially supported by the Wallenberg AI, Autonomous Systems and 
Software Program (WASP) funded by the Knut and Alice Wallenberg Foundation.
Computing resources were provided by the Swedish National Infrastructure for 
Computing (SNIC) and the Swedish National Supercomputer Centre (NSC).

\bibliographystyle{splncs04}
\bibliography{refs.bib}

\begin{thebibliography}{10}
\providecommand{\url}[1]{\texttt{#1}}
\providecommand{\urlprefix}{URL }
\providecommand{\doi}[1]{https://doi.org/#1}

\bibitem{Adadi18}
Adadi, A., Berrada, M.: Peeking inside the black-box: A survey on explainable
  artificial intelligence ({XAI}). IEEE access  \textbf{6},  52138--52160
  (2018)

\bibitem{Arrieta20}
{Barredo Arrieta}, A., Díaz-Rodríguez, N., {Del Ser}, J., Bennetot, A.,
  Tabik, S., Barbado, A., Garcia, S., Gil-Lopez, S., Molina, D., Benjamins, R.,
  Chatila, R., Herrera, F.: Explainable artificial intelligence ({XAI}):
  Concepts, taxonomies, opportunities and challenges toward responsible {AI}.
  Information Fusion  \textbf{58},  82 -- 115 (2020)

\bibitem{Bartocci19}
Bartocci, E., Beyer, D., Black, P.E., Fedyukovich, G., Garavel, H., Hartmanns,
  A., Huisman, M., Kordon, F., Nagele, J., Sighireanu, M., et~al.: Toolympics
  2019: An overview of competitions in formal methods. In: International
  Conference on Tools and Algorithms for the Construction and Analysis of
  Systems. Springer (2019)

\bibitem{Breiman01}
Breiman, L.: Random forests. Machine learning  \textbf{45}(1),  5--32 (2001)

\bibitem{Cousot77}
Cousot, P., Cousot, R.: Abstract interpretation: A unified lattice model for
  static analysis of programs by construction or approximation of fixpoints.
  In: Conference Record of the Fourth Annual ACM SIGPLAN-SIGACT Symposium on
  Principles of Programming Languages. pp. 238--252. ACM Press, New York, NY
  (1977)

\bibitem{Devos21}
Devos, L., Meert, W., Davis, J.: Verifying tree ensembles by reasoning about
  potential instances. In: SIAM International Conference on Data Mining
  proceedings. SIAM (2021)

\bibitem{Einziger19}
Einziger, G., Goldstein, M., Sa’ar, Y., Segall, I.: Verifying robustness of
  gradient boosted models. In: Proceedings of the Thirty-Third {AAAI}
  Conference on Artificial Intelligence. pp. 2446--2453. {AAAI} Press (2019)

\bibitem{Friedman01}
Friedman, J.H.: Greedy function approximation: A gradient boosting machine.
  Annals of statistics pp. 1189--1232 (2001)

\bibitem{Julian19b}
Julian, K.D., Kochenderfer, M.J.: Guaranteeing safety for neural network-based
  aircraft collision avoidance systems. In: 2019 IEEE/AIAA 38th Digital
  Avionics Systems Conference (DASC). pp. 1--10. IEEE (2019)

\bibitem{Julian19a}
Julian, K.D., Kochenderfer, M.J., Owen, M.P.: Deep neural network compression
  for aircraft collision avoidance systems. Journal of Guidance, Control, and
  Dynamics  \textbf{42}(3),  598--608 (2019)

\bibitem{Katz17}
Katz, G., Barrett, C., Dill, D.L., Julian, K., Kochenderfer, M.J.: Reluplex: An
  efficient {SMT} solver for verifying deep neural networks. In: International
  Conference on Computer Aided Verification (CAV). pp. 97--117. Springer
  International Publishing (2017)

\bibitem{MNIST}
LeCun, Y., Bottou, L., Bengio, Y., Haffner, P.: Gradient-based learning applied
  to document recognition. Proceedings of the IEEE  \textbf{86}(11),
  2278--2324 (1998)

\bibitem{Liu21}
Liu, C., Arnon, T., Lazarus, C., Strong, C., Barrett, C., Kochenderfer, M.J.:
  Algorithms for verifying deep neural networks. Foundations and Trends in
  Optimization  \textbf{4}(3-4),  244--404 (2021)

\bibitem{Nicolae18}
Nicolae, M.I., Sinn, M., Tran, M.N., Buesser, B., Rawat, A., Wistuba, M.,
  Zantedeschi, V., Baracaldo, N., Chen, B., Ludwig, H., et~al.: Adversarial
  robustness toolbox v1.0.0. arXiv preprint arXiv:1807.01069  (2018)

\bibitem{Pedregosa11}
Pedregosa, F., Varoquaux, G., Gramfort, A., Michel, V., Thirion, B., Grisel,
  O., Blondel, M., Prettenhofer, P., Weiss, R., Dubourg, V., et~al.:
  Scikit-learn: Machine learning in {P}ython. Journal of machine learning
  research  \textbf{12},  2825--2830 (2011)

\bibitem{Prokhorenkova18}
Prokhorenkova, L., Gusev, G., Vorobev, A., Dorogush, A.V., Gulin, A.: Catboost:
  Unbiased boosting with categorical features. In: Advances in Neural
  Information Processing Systems (NIPS) (2018)

\bibitem{Ranzato20}
Ranzato, F., Zanella, M.: Abstract interpretation of decision tree ensemble
  classifiers. In: Proceedings of the Thirty-Fourth {AAAI} Conference on
  Artificial Intelligence. pp. 5478--5486. {AAAI} Press (2020)

\bibitem{Tjoa20}
Tjoa, E., Guan, C.: A survey on explainable artificial intelligence ({XAI}):
  Toward medical {XAI}. IEEE Transactions on Neural Networks and Learning
  Systems  (2020)

\bibitem{Tornblom19}
T{\"o}rnblom, J., Nadjm-Tehrani, S.: An abstraction-refinement approach to
  formal verification of tree ensembles. In: International Conference on
  Computer Safety, Reliability, and Security (SAFECOMP). pp. 301--313. Springer
  (2019)

\bibitem{Tornblom20}
T{\"o}rnblom, J., Nadjm-Tehrani, S.: Formal verification of input-output
  mappings of tree ensembles. Science of Computer Programming  \textbf{194}
  (2020)

\end{thebibliography}

\end{document}